%% file: main.tex
\documentclass[10pt,twocolumn,letterpaper]{article}

\usepackage{iccv}
\usepackage{times}
\usepackage{epsfig}
\usepackage{graphicx}
\usepackage{amsmath}
\usepackage{amssymb}
\usepackage{xcolor}
\usepackage{verbatim}
\usepackage{caption}
\usepackage{enumitem}
\usepackage{wrapfig}
\usepackage{appendix}

\usepackage[pagebackref=true,breaklinks=true,letterpaper=true,colorlinks,bookmarks=false]{hyperref}

\iccvfinalcopy 

\def\iccvPaperID{****} 

\input{symbols_format.tex}

\makeatletter
\renewcommand{\paragraph}{%
  \@startsection{paragraph}{4}%
  {\z@}{0.65ex \@plus 1ex \@minus .2ex}{-0.75em}%
  {\normalfont\normalsize\bfseries}%
}
\makeatother

\ificcvfinal\fi

\begin{document}

\title{Neural Strokes: Stylized Line Drawing of 3D Shapes}

\author{ 
        Difan Liu$^{1}$
    \and
        Matthew Fisher$^2$
    \and
        Aaron Hertzmann$^2$
    \and
        Evangelos Kalogerakis$^1$
     \and \vspace{-5mm}
     \\ 
    $^1$University of Massachusetts Amherst \,\,\,\,\,\,\,\,\,\,\,\,\,\,\, $^2$Adobe Research
}

\twocolumn[{%
 \renewcommand\twocolumn[1][]{#1}%
 \maketitle
 \vspace{-6mm}
 \centering
\includegraphics[width=\textwidth]{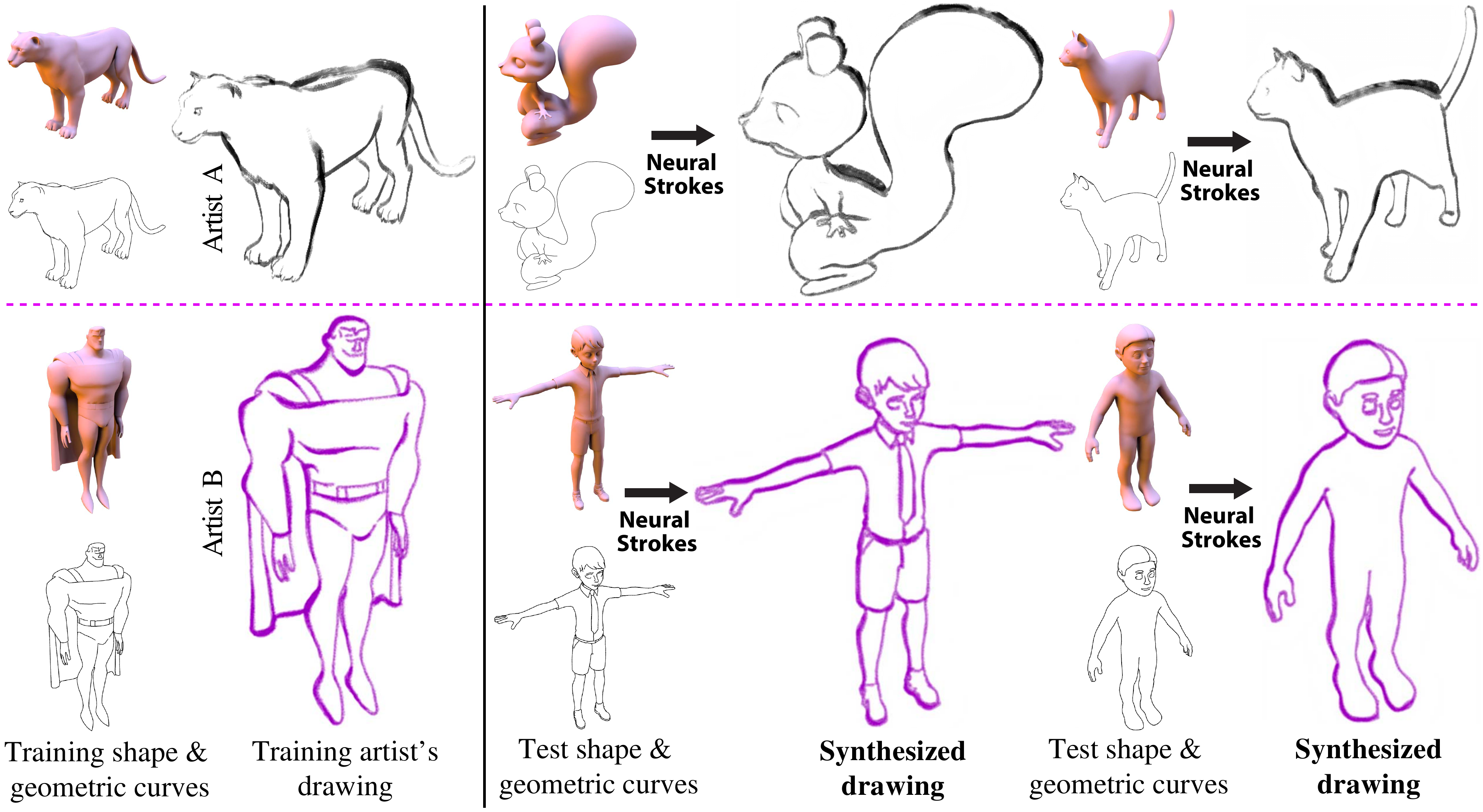}
 \vspace{-7mm}
 \captionof{figure}
  { 
    Our model learns to generate stylized line drawings from a single example of a training  shape and corresponding  drawing. Given a test 3D shape and 2D geometric curves representing the shape, our model synthesizes a line drawing in the style of the training example. Here we show  synthesized drawings by transferring the artist's style A (top) or B (below).
    \label{fig:teaser}
   }   
\vspace{4mm}   
}]

\ificcvfinal\thispagestyle{empty}\fi

\begin{abstract}
\vspace{-3mm}
This paper introduces a model for producing stylized line drawings from 3D shapes.  The model takes a 3D shape and a viewpoint as input, and outputs a drawing with textured strokes, with variations in stroke thickness, deformation, and color learned from an artist's style. The model is fully differentiable. We train its parameters from a single training drawing of another 3D shape.
We show that, in contrast to previous image-based methods, the use of a geometric representation of 3D shape and 2D strokes allows the model to transfer important aspects of shape and texture style while preserving contours. Our method outputs the resulting drawing in a vector representation, enabling richer downstream analysis or editing in interactive applications. Our code and dataset are available at our project page:
\mbox{\url{www.github.com/DifanLiu/NeuralStrokes}}

\end{abstract}

\section{Introduction}
\input{./tex/intro}

\section{Related Work}
\input{./tex/related_work}

\section{Model}
\input{./tex/method}

\section{Training}
\input{./tex/training}

\section{Experiments}

\input{./tex/result}

\section{Conclusion}
\input{./tex/conclusion}

{\small
\bibliographystyle{ieee_fullname}
\bibliography{egbib,perception,contour_tutorial}
}

\clearpage
\renewcommand\appendixpagename{Appendix}
\appendix
\appendixpage

\subsection*{1. Additional comparisons and results}

First, we note that our source code and dataset are available on our project website:\\
\url{www.github.com/DifanLiu/NeuralStrokes}

\begin{figure}[h!]
\begin{center}
\includegraphics[width=\linewidth]{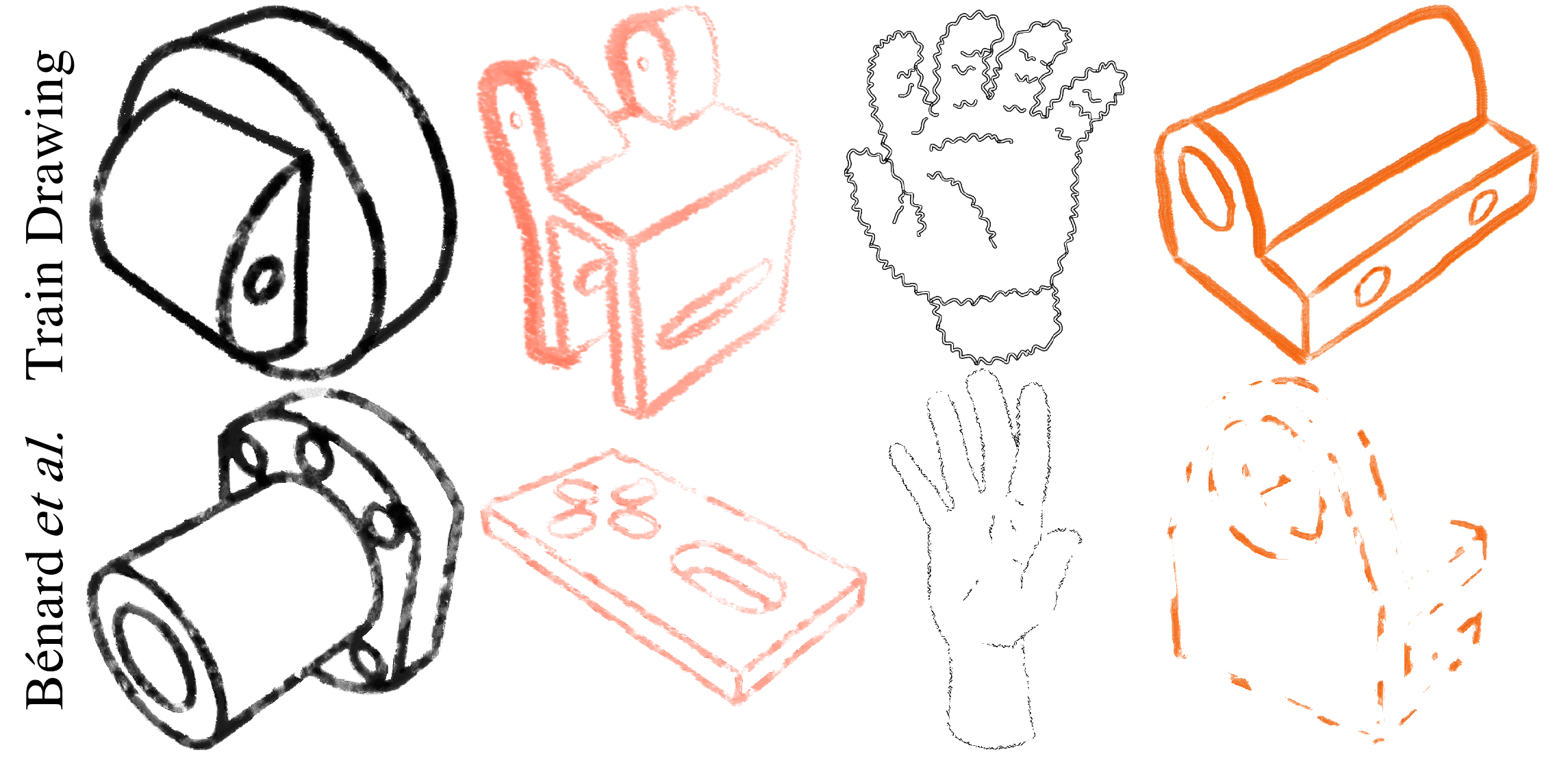}
\end{center}
\vspace{-5mm}
   \caption{\emph{Top:} artist-drawn training drawings. \emph{Bottom:} results from B{\'e}nard \etal \cite{Benard:2013}.}
\vspace{-2mm}
\label{fig:PaintTween}
\end{figure}

\paragraph{Additional comparisons with B{\'e}nard \etal \cite{Benard:2013}.}
Figure \ref{fig:PaintTween} shows the training artist’s drawing on the top, and results from B{\'e}nard \etal \cite{Benard:2013} in the bottom (zoom-in for details, and compare with our results in Fig. 4, 8, 5 of our main paper). They roughly capture the overall distribution of line properties, without matching the artist’s choices well. Moreover, B{\'e}nard \etal \cite{Benard:2013} introduces many holes and cannot handle challenging cases, such as varying stroke thickness or large deformation (the rightmost style in Figure \ref{fig:PaintTween}).

\begin{figure}[t]
\begin{center}
\includegraphics[width=\linewidth]{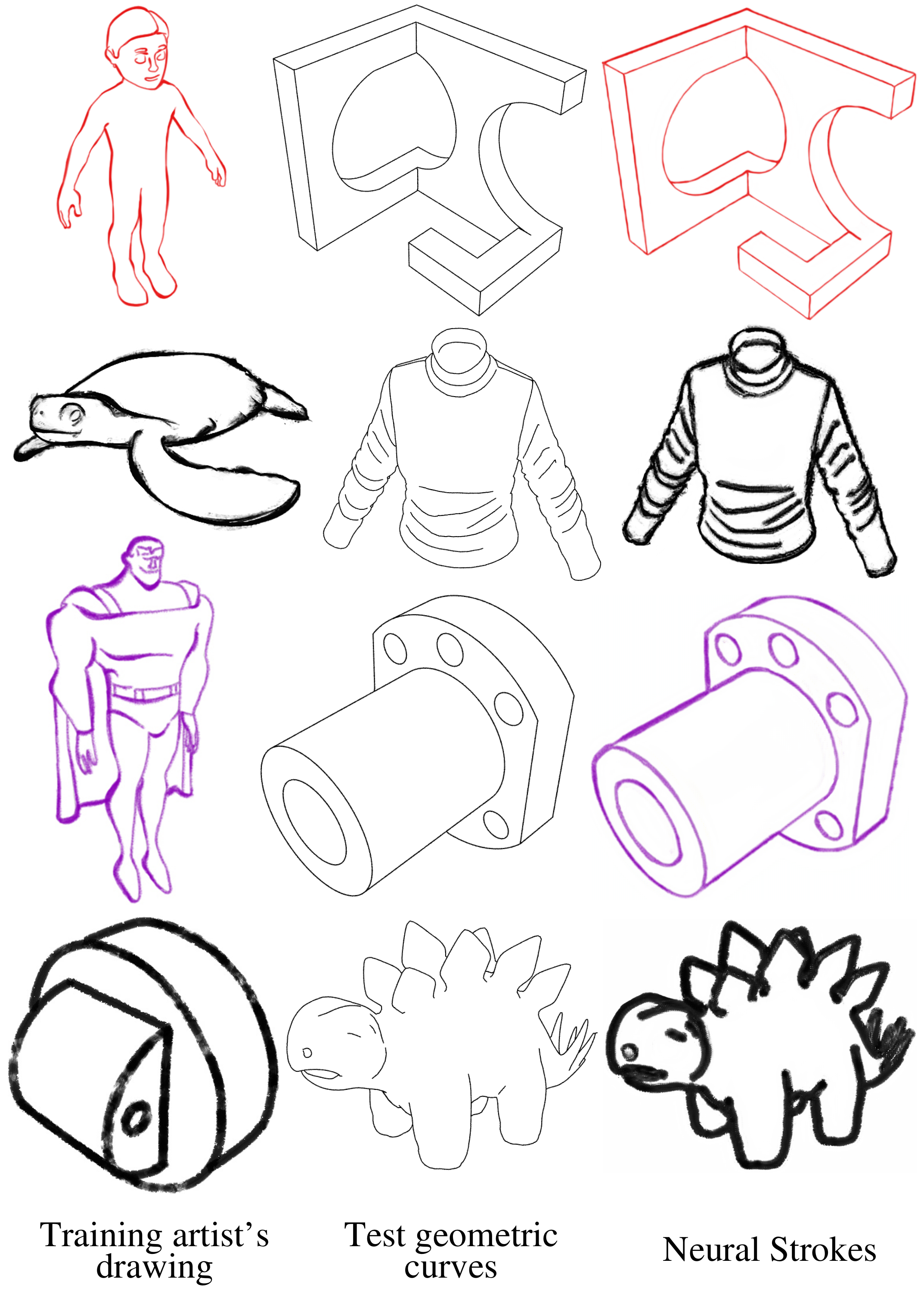}
\end{center}
\vspace{-4mm}
   \caption{\emph{Left to right:} training artist's drawing, test geometric curves, Neural Strokes.}
\label{fig:inter_class}
\end{figure}

\paragraph{More generalization cases.}
Figure \ref{fig:inter_class} demonstrates challenging generalization cases: given 
a training drawing of a shape belonging to one category (e.g., humanoid), we synthesize a drawing for a shape from an entirely different category (e.g., mechanical object) in the same style. Our method still generalizes sufficiently in these challenging cases.

\subsection*{2. Network architecture}
We provide here additional details of our network architecture (see also Section 3.2 and 3.3 of our main text).

\paragraph{Surface geometry module.} Our surface geometry module uses the architecture shown in Table \ref{table:arch_surface}. All convolutional layers are followed by instance normalization \cite{instancenorm} and a ReLU nonlinearity. The module contains $4$  residual blocks \cite{he2016deep}, where each residual block contains two $3 \times 3$ convolutional layers with the same number of filters for both layers. 

\begin{table}[h!]
\begin{center}
\begin{tabular}{|c|c|}
\hline
\textbf{Layer} & \textbf{Activation size}  \\
\hline\hline
Input & 768 $\times$ 768 $\times$ 9  \\
Conv2D(7x7, 9$\rightarrow$10, stride=1)   & 768 $\times$ 768 $\times$ 10  \\
Conv2D(3x3, 10$\rightarrow$20, stride=2)   & 384 $\times$ 384 $\times$ 20  \\
Conv2D(3x3, 20$\rightarrow$40, stride=2)   & 192 $\times$ 192 $\times$ 40  \\
4 Residual blocks   & 192 $\times$ 192 $\times$ 40  \\
Conv2D(3x3, 40$\rightarrow$40, stride=1/2)   & 384 $\times$ 384 $\times$ 40  \\
Conv2D(3x3, 40$\rightarrow$40, stride=1/2)   & 768 $\times$ 768 $\times$ 40  \\
Conv2D(1x1, 40$\rightarrow$40, stride=1)   & 768 $\times$ 768 $\times$ 40  \\
\hline
\end{tabular}
\end{center}
\caption{Architecture of the surface geometry module.}
\label{table:arch_surface}
\end{table}

\paragraph{Path geometry module.} Our path geometry module uses the architecture shown in Table \ref{table:arch_path}. The first two convolutional layers are followed by a ReLU nonlinearity. The last layer has 3 output channels: two for 2D displacement, and one for thickness. For thickness, we use a ReLU activation to guarantee non-negative outputs, while for the 2D real-valued displacement output, we do not use any nonlinearity.
\begin{table}[h!]
\begin{center}
\begin{tabular}{|c|c|}
\hline
\textbf{Layer} & \textbf{Activation size}  \\
\hline\hline
Input & $M_i$ $\times$ 45  \\
Conv1D(3x3, 45$\rightarrow$40, stride=1)   & $M_i$ $\times$ 40  \\
Conv1D(3x3, 40$\rightarrow$40, stride=1)   & $M_i$ $\times$ 40  \\
Conv1D(3x3, 40$\rightarrow$3, stride=1)   & $M_i$ $\times$ 3  \\
\hline
\end{tabular}
\end{center}
\caption{Architecture of the path geometry module.}
\label{table:arch_path}
\end{table}

\paragraph{Stroke texture module.} Our stroke texture module uses the architecture shown in Table \ref{table:arch_texture}. All convolutional layers are followed by instance normalization \cite{instancenorm} and a ReLU nonlinearity except for the last convolutional layer. The last convolutional layer is followed by a sigmoid activation function. The module contains $6$ residual blocks \cite{he2016deep}, where each residual block contains two $3 \times 3$ convolutional layers with the same number of filters for both layers. 

\begin{table}[t!]
\begin{center}
\begin{tabular}{|c|c|}
\hline
\textbf{Layer} & \textbf{Activation size}  \\
\hline\hline
Input & 768 $\times$ 768 $\times$ 9  \\
Conv2D(7x7, 9$\rightarrow$64, stride=1)   & 768 $\times$ 768 $\times$ 64  \\
Conv2D(3x3, 64$\rightarrow$128, stride=2)   & 384 $\times$ 384 $\times$ 128  \\
Conv2D(3x3, 128$\rightarrow$256, stride=2)   & 192 $\times$ 192 $\times$ 256  \\
6 Residual blocks   & 192 $\times$ 192 $\times$ 256  \\
Conv2D(3x3, 256$\rightarrow$128, stride=1/2)   & 384 $\times$ 384 $\times$ 128  \\
Conv2D(3x3, 128$\rightarrow$64, stride=1/2)   & 768 $\times$ 768 $\times$ 64  \\
Conv2D(7x7, 64$\rightarrow$3, stride=1)   & 768 $\times$ 768 $\times$ 3  \\
\hline
\end{tabular}
\end{center}
\caption{Architecture of the stroke texture module.}
\label{table:arch_texture}
\end{table}

\subsection*{3. Additional experiments}
\begin{figure}[t!]
\begin{center}
\includegraphics[width=\linewidth]{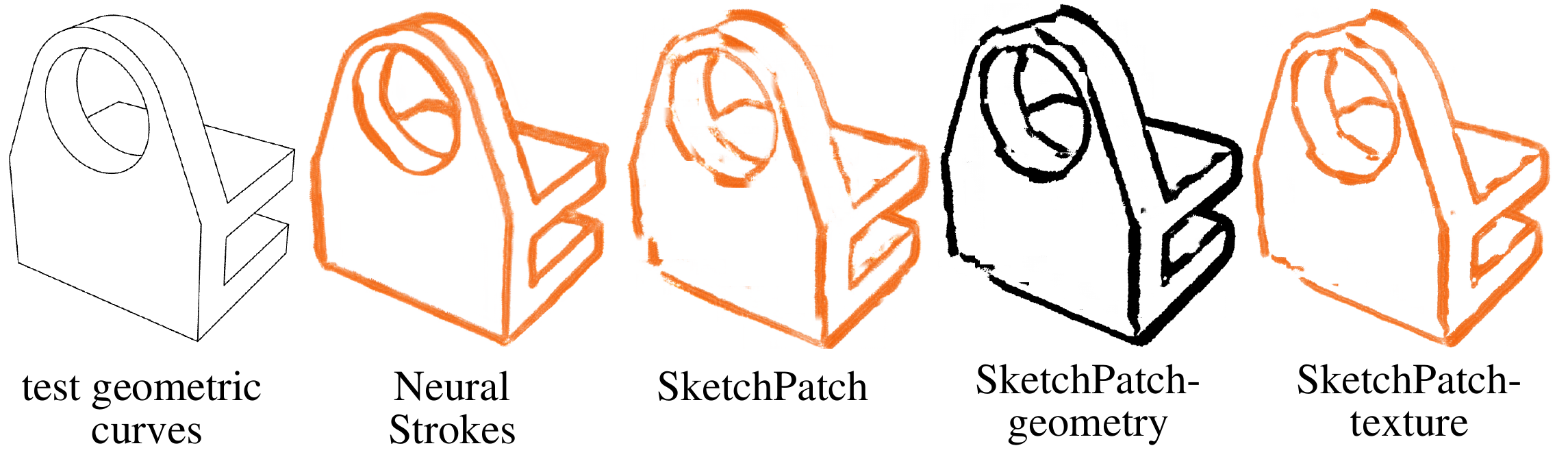}
\end{center}

   \caption{\emph{Left to right:} test geometric curves, Neural Strokes, SketchPatch, SketchPatch-geometry, SketchPatch-texture result.}

\label{fig:supp_exp}
\end{figure}

We experimented with using one SketchPatch model for stroke geometry prediction and another SketchPatch model for stroke texture prediction, as discussed in Section 5 of our main text (``comparison methods'' paragraph).  Specifically, in the first step, we train a SketchPatch model (called \emph{SketchPatch-geometry}) on the training stroke mask $\hat{\bI}_b$ to predict stroke geometry as a grayscale raster image. In the second step, we train another SketchPatch model (called \emph{SketchPatch-texture}) on the training drawing $\hat{\bI}$ to generate a stylized line drawing given the output of \emph{SketchPatch-geometry}. 
The results did not improve compared to \emph{SketchPatch} in terms of our evaluation metrics (see Table \ref{table:supp_exp_table}). Figure \ref{fig:supp_exp} shows example output of \emph{SketchPatch-geometry} and \emph{SketchPatch-texture}. 

\begin{table}[t!]
\begin{center}
\begin{tabular}{|c|c|c|}
\hline
Method & LPIPS $\downarrow$ & FID $\downarrow$ \\
\hline\hline
\emph{SketchPatch} & 0.1104 & 83.60  \\
\emph{SketchPatch-texture} & 0.1142 & 86.96  \\
\cline{1-3}
\emph{Neural Strokes}& \textbf{0.0956} & \textbf{62.40}   \\
\hline
\end{tabular}
\end{center}

\caption{Quantitative evaluation of SketchPatch variants.}
\label{table:supp_exp_table}
\end{table}

\begin{figure}[t!]
\begin{center}
\includegraphics[width=\linewidth]{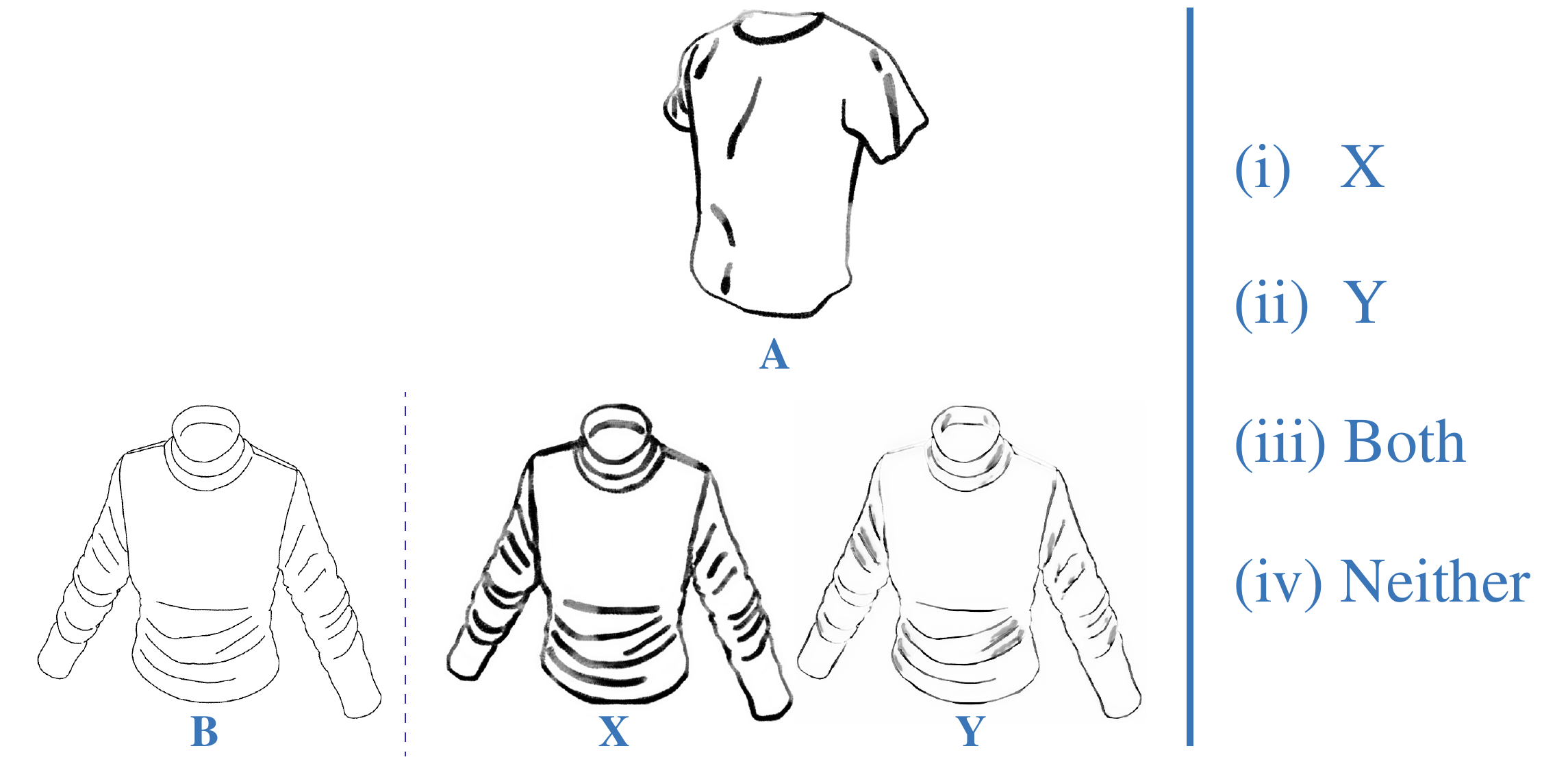}
\end{center}

   \caption{Layout shown to participants of our user study.}
\vspace{-4mm}
\label{fig:query_layout}
\end{figure}
\subsection*{4. Perceptual evaluation}
We conducted an Amazon Mechanical Turk perceptual evaluation where we showed participants (a) a stylized artist’s drawing for a training shape (Figure \ref{fig:query_layout}, A)
, (b) test geometric curves
(Figure \ref{fig:query_layout}, B)
, 
(c) a pair of stylized line drawings of the test shape placed in a randomized left/right position (Figure \ref{fig:query_layout}, X and Y): one line drawing was picked from our method, while the other came from \emph{SketchPatch},  \emph{SinCUT}, \emph{NST}, B{\'e}nard \etal \cite{Benard:2013}, or \emph{Artists} ($5$ possible comparison cases).  We asked participants to select the drawing that best mimicked the style of training drawing A. Participants could pick one of four options: drawing X, drawing Y, ``neither of the drawings mimicked the style well'', or ``both drawings mimicked the style well''. The study included the $31$ styles from our dataset and each style consists of $3$ test shapes. As a result, there were total $93$ test cases, each involving the above-mentioned $5$ comparisons ($465$ total comparisons).

Each questionnaire was released via the MTurk platform. It contained $15$ unique questions, each asking for one comparison. Then these $15$ questions were repeated in the questionnaire in a random order. In these repeated questions, the order of compared line drawings was flipped. If a worker gave more than $5$ inconsistent answers for the repeated questions, then the worker was marked as ``unreliable''. Each participant was allowed to perform the questionnaire only once to ensure participant diversity. A total of $161$ participants took part in the study. Among $161$ participants, $68$ workers were marked as ``unreliable''. For each of the $465$ comparisons, we gathered votes from $3$ different ``reliable'' users. The results are shown in Figure 6 of the main text.

\end{document}

%% file: symbols_format.tex
\newcommand{\bc}{\mathbf{c}}
\newcommand{\bd}{\mathbf{d}}
\newcommand{\be}{\mathbf{e}}

\newcommand{\bn}{\mathbf{n}}

\newcommand{\bt}{\mathbf{t}}

\newcommand{\bw}{\mathbf{w}}

\newcommand{\bC}{\mathbf{C}}

\newcommand{\bF}{\mathbf{F}}

\newcommand{\bI}{\mathbf{I}}

\newcommand{\bP}{\mathbf{P}}

\newcommand{\bU}{\mathbf{U}}
\newcommand{\bV}{\mathbf{V}}

\def\D{{\cal D}}

\makeatletter
\renewcommand{\paragraph}{%
  \@startsection{paragraph}{4}%
  {\z@}{1ex \@plus 1ex \@minus .2ex}{-1em}%
  {\normalfont\normalsize\bfseries}%
}

%% file: tex/intro.tex
\vspace{-2mm}
Understanding and creating stylized outline drawings is a key task for stylization \cite{Liu:2020:NC,BenardHertzmann}, sketch understanding \cite{Lift3D_SA20}, and human vision \cite{Cole:2009,Hertzmann:2020:WDL}.  Artists and amateurs alike draw pictures of 3D objects in many different styles, whether for art, animation, architectural design, 3D authoring, or simply the pleasure of drawing.  However, most recent research in image stylization does not take 3D geometry into account, producing drawings that frequently lose detail and do not capture image outlines. Conversely, there is a long history of 3D drawing algorithms that create precise line drawings in hand-authored procedural styles, but they cannot be learned from data, making them difficult to control and inapplicable for analysis of existing sketches.  While there is a long literature on analysis and shape reconstruction from sketches, these methods typically assume that artists draw with plain line styles.

This paper introduces differentiable learning for stylization of 3D shapes, combining ideas from classic 3D line drawing algorithms with modern differentiable rendering. The algorithm produces a stylized vector rendering from a 3D shape, in a style learned from a single drawing made by an artist of a reference shape (Figure \ref{fig:teaser}). There are several challenges in making such a system work. To generate stylized strokes, our method must 
disentangle several stroke attributes, including spatially-varying thickness, geometric deformations and smoothness, and texture.  These elements may often be quite noisy, with strokes being wiggly or messy; the final pixel values are an entangled combination of these factors. Although we train from a drawing paired with a 3D shape, the individual components of thickness, deformation and texture are not labeled in the data. Training with a purely pixel-based image translation model fails to disentangle these factors, producing noisy results that lose image details.  Moreover, due to the difficulties of producing skilled artistic drawings, our training set is necessarily  small, with only one drawing per distinct style.

To address these challenges, we propose a differentiable rendering formulation of stroke attributes. This allows the model to learn to accurately predict stroke thickness, deformation, and texture. Because our method works with 3D geometry, it can accurately capture fine details of a shape that purely image-based methods cannot. Our model is trained on  pixel inputs, yet it produces output in a vector graphics format, which provides the benefits of a compact representation with infinite resolution and is suitable for further editing and use by graphic designers. 

Our technical contributions include: (a) a convolutional network operating along parameterized stroke paths, (b)  the combination of 3D geometry, 2D image, and 1D curve feature maps to learn stroke properties with a differentiable vector renderer, and (c) learning from a single example based on multi-scale patches of the training drawing. We show that our method produces substantially better results than existing image-based methods, in terms of predicting artists' drawings, and in user evaluation of results.

%% file: tex/related_work.tex
Many methods learn to stylize images from training examples. The first such methods, Image Analogies \cite{Hertzmann:2001} and Neural Style Transfer \cite{gatys2016image},
used only single-image style exemplars. Many variants of Neural Style Transfer use Gram-matrix-like losses for training or optimization from single examples, e.g., \cite{Johnson2016Perceptual,adain,wct}.  Other recent approaches learn stylization from larger collections of paired \cite{isola2017image,im2pencil} or unpaired examples \cite{zhu2017unpaired,cut}. All of these methods take only images for input and output. However, these methods lose important geometric information, often resulting in inaccurate portrayal of shape, such as broken outlines. Moreover, these methods do not produce vector output, limiting their usefulness for certain applications.

Stylized rendering of 3D shapes has a long history in Non-Photorealistic Rendering (NPR) research \cite{BenardHertzmann}, and these algorithms have been used in numerous applications, including movies  \cite{Coleman:2020}, and video games  \cite{Thibault:2010}.  
Most methods entail hand-designed procedural stylization, e.g., \cite{Grabli:2010,Winkenbach:1994,Winkenbach:1996,Kalogerakis:2009:ddcurvature}. None of these methods can learn stylization from examples, making authoring and definition of styles challenging.  Moreover, none of these methods are differentiable, making them unsuitable for integration with other vision tasks, such as sketch analysis and interpretation.

A few previous methods learn stylized 3D rendering. B{\'e}nard \etal \cite{Benard:2013} and StyLit \cite{Fiser:2016} extend Image Analogies to stylize 3D models and animation. Neural Contours~\cite{Liu:2020:NC} learns to select which outlines to draw. Our work is complementary, since we learn to stylize outline curves. Moreover, our method produces vector rather than raster output, which is a more interpretable and useful representation.

Our work builds on ideas from learning vector strokes and stylization. Most existing methods for example-based stroke stylization \cite{Hertzmann:2002, lu2012helpinghand, lu2013realbrush, Kalnins:2002} are not differentiable and require vector training data. More recent methods define differentiable strokes for painting and vector graphics \cite{ganin2018synthesizing, li2020differentiable}, though these methods do not support texture synthesis.  Our work is perhaps most similar to SketchPatch \cite{fish2020sketchpatch}. SketchPatch is an image-to-image model that translates a plain sketch to a textured sketch. However, SketchPatch does not take 3D shape and stroke geometry into consideration, and its output is a raster image, and as a result, the method often loses detail from the input geometry.

\begin{figure*}[t!]
\begin{center}
\includegraphics[width=\textwidth]{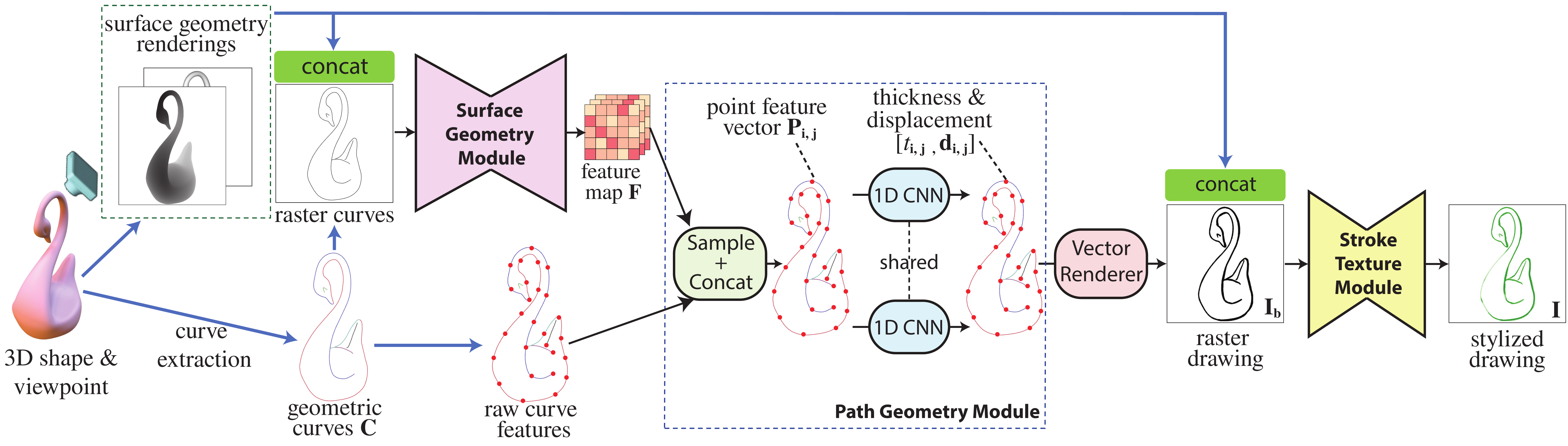}
\end{center}
\vspace{-5mm}
   \caption{Our network architecture: the input 3D shape and a set of geometric curves are processed by a surface geometry module and a path geometry module to produce stroke thickness and displacement. With the predicted thickness and displacement, a stroke texture module creates a stylized line drawing with texture.}
\vspace{-2mm}   
\label{fig:architecture}
\end{figure*}

%% file: tex/method.tex
 \label{sec:method}

This section describes our stylized rendering architecture. The subsequent section describes how to train the model from a single artist-drawn example. 
Our trained model (Figure \ref{fig:architecture}) takes as input a 3D shape and a camera position and produces a stylized vector rendering $\bI$, represented as a set of strokes, that is, curves with varying thickness and texture.
This allows us to simulate the appearance of drawing media (pen, pencil, paint), and the ways artists vary pressure/thickness along strokes \cite{Guptill:1997,Hodges:2003}.

\subsection{Curve Extraction}
 \label{sec:curve_extraction}
The first stage of our model extracts geometric curves from the 3D shape, producing a set of  curves $\bC$. The goal of the rest of our model is to convert these plain curves into stylized strokes, by assigning thickness, displacement, and texture along these curves.

The curves are extracted using  an existing algorithm for creating line drawings from 3D shapes.  Many such methods have been developed \cite{DeCarlo:2012}, and our method can be used with any method that outputs vector curves. In this paper, we extract curves using the pretrained Geometry Branch of Neural Contours \cite{Liu:2020:NC}, which combines curves from many prior algorithms, including Occluding Contours \cite{BenardHertzmann}, Suggestive Contours \cite{DeCarlo:2003}, Apparent Ridges \cite{Judd:2007}, Ridges \& Valleys \cite{Ohtake:2004}.

The geometric curves are represented as polylines: $N$ vector paths $\bC=\{\bc_i\}_{i=1}^N$, where $\bc_i$ is  a sequence of densely sampled points $\bc_i=\{\bc_{i,j}\}_{j=1}^{M_i}$ with uniform spacing, $M_i$ is the number of points on the path, and $\bc_{i,j}$ represents the 2D position of point $j$ on  path $i$. 

\subsection{Stroke geometry prediction}

The central portion of our model, described in this section, produces one stylized output stroke for each of the curves in $\bC$.  Texture synthesis is described in Section \ref{sec:texture_module}.

\begin{figure}[t]
\begin{center}
\includegraphics[width=\linewidth]{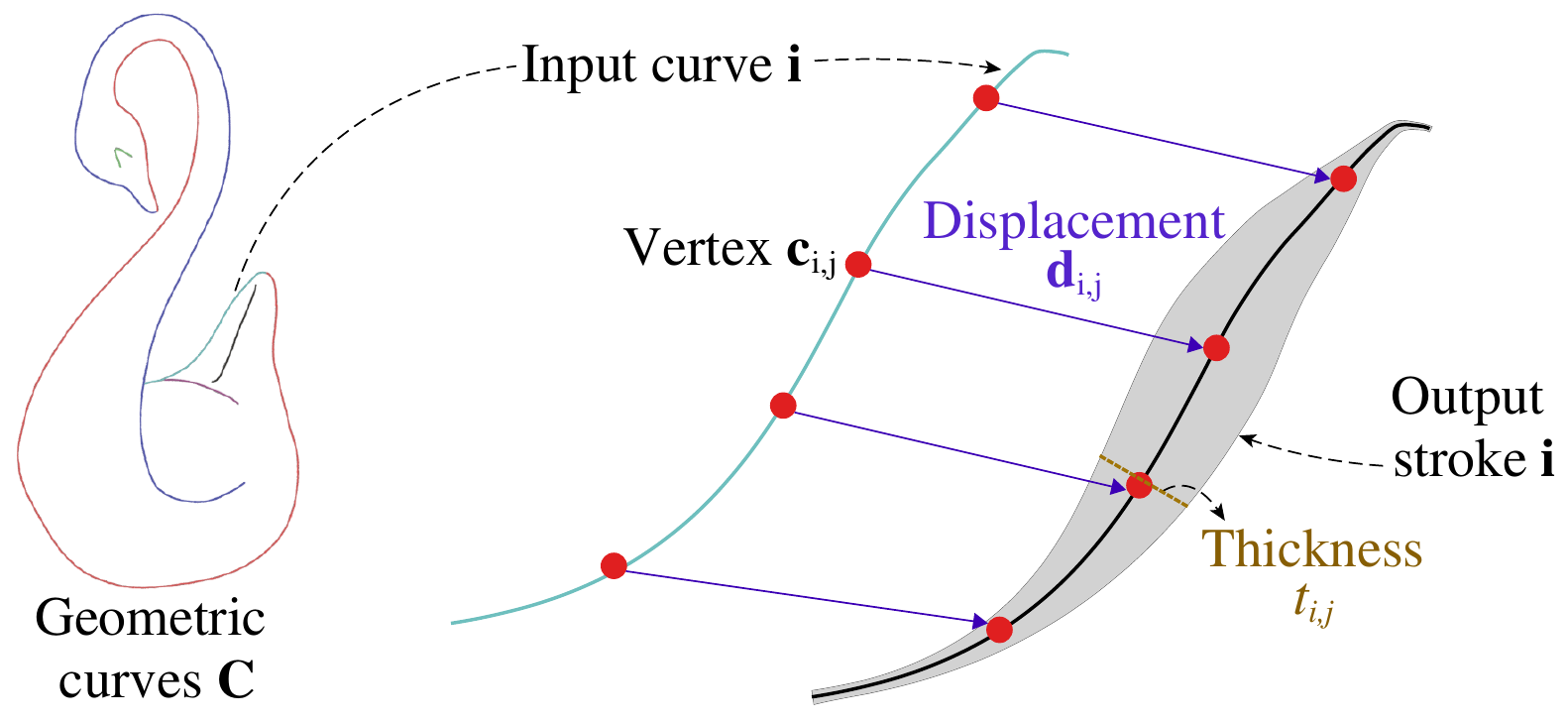}
\end{center}
\vspace{-6mm} 
   \caption{\emph{Left:} an input set of geometric curves (each curve is highlighted with a different color). \emph{Right:} For each input curve, our model outputs a stroke by predicting a thickness scalar and a 2D displacement vector for each control point.}
\vspace{-3mm}    
\label{fig:data_structure}
\end{figure}

The geometric curves $\bC$ are unstylized, and the goal of our model is to convert them to \textit{strokes} with new shape, along with thickness and texture. The stroke control points are represented as displacements from the input curves. Displacement models the ways that artists deform curves, for example, smoothing curves, adding ``wiggles'' (e.g. Figure \ref{fig:intermediate}), and so on \cite{Fankbonner:2003}.
More precisely, for each input polyline $\bc_i$, the module described in this section produces a 1D thickness $t_{i,j}$ for each control point $\bc_{i,j}$, together with a displacement vector $\bd_{i,j}$. Hence, the control points of the output stroke will be $\{ \bc_{i,j} + \bd_{i,j} \}$ (Figure \ref{fig:data_structure}).

As observed in previous work, artistic stroke thickness and displacement depend on image-space geometric shape features (e.g., object depth, view-dependent curvature, and surface shading \cite{Goodwin:2007}). Stroke thickness and displacement also depend on the shape of the stroke itself, including phenomena like tapering, stroke smoothing, and ``wiggliness,'' (e.g., \cite{Grabli:2010}), which can be captured as deformations of the 1D curve.  Hence, to predict stroke geometry, the model includes modules to incorporate information from both the shape's surface geometry and along the 1D stroke paths.

\label{sec:geometry_model}

\paragraph{Surface geometry module.}
First, the surface geometry module processes surface geometry via a 2D convolutional neural network, outputting image-space feature maps $\bF$.

Surface geometry is represented in the form of image-space renderings.  Each pixel contains the geometric properties of the surface point that projects to that pixel. There are nine input channels per pixel: depth from camera, radial curvature, derivative of radial curvature \cite{DeCarlo:2003}, maximum and minimum principal surface curvatures, view-dependent surface curvature \cite{Judd:2007}, dot product of surface normal with view vector,
and a raster image containing the line segments of the vector paths $\bC$. 
We found these geometric and shading features to be useful for predicting accurate stroke geometry. 
In this manner, the module jointly processes shape features and vector paths in the concatenated $768\times768\times9$ map $\bV$. The map passes through a  neural network function to output a  $768\times768\times40$  deep feature map $\bF = f(\bV;\bw_1)$, where $f$ is a ResNet-based fully convolutional network \cite{Johnson2016Perceptual} with four residual blocks, and $\bw_1$ are learned during training. 
More details about the module architecture are given in the appendix.
   
\paragraph{Path geometry module.} The path geometry module is a neural network applied separately to each input curve using 1D convolutions. Each point $\{i,j\}$ on a curve has a set of curve features and features from the shape geometry.  

The curve features are 2D curve normals, 2D tangent directions, and the normalized arc length. The normalized arc length allows the model to learn to taper stroke thickness, whereas the other two features can capture image-space curve orientations. 
Since the orientation of the curve is ambiguous,  there is a sign ambiguity in the tangent direction $\be_{i,j}$ and normal $\bn_{i,j}$ per curve point. To handle the ambiguity, we extract two alternative curve features sets: one using $(\be_{i,j},\bn_{i,j})$ and another set using $(-\be_{i,j},-\bn_{i,j})$. 

In addition to the curve features, the deep surface geometry features $\bF$ generated by the surface geometry module are also included as input to the path geometry module. Specifically, for each point on a curve, we use nearest interpolation on the deep feature map $\bF$ to produce $40$-dim  features. These features are concatenated with each of the two sets of the above $5$ raw curve features of the vector path, resulting in 
two $M_i \times 45$ feature  maps $(\bP_i,\bP'_i)$ for the path $i$, where $M_i$ is the  number of control points in the path.
In this manner, the module jointly processes view-based surface features  together with geometric properties specific to the path.
We also experimented with processing these features independently and found the above combination yielded the best performance.
The above features pass through a  neural network function to predict the thickness and 2D displacement along each vector path:
\begin{equation}
[\bt_i, \bd_i] = 
\mathrm{avg}\big( h(\bP_i;\bw_2), h(\bP'_i;\bw_2) \big)
\end{equation}
where  $\bd_i=\{\bd_{i,j}\}_{j=1}^{M_i}$ are the predicted per-point displacements, and $\bt_i=\{t_{i,j}\}_{j=1}^{M_i}$ are   per-point thicknesses (Figure \ref{fig:data_structure}), and
$\bw_2$ are parameters learned from the training reference drawing. The $\mathrm{avg}$ function performs average pooling over predictions of the two alternative feature sets to ensure invariance to the sign of curve orientation. 
The function
 $h$ is a 1D fully convolutional network made of $3$ layers, each using filters of kernel size $3$, stride $1$, zero padding. The first two layers are followed by ReLU activation. 
 The last layer has $3$ output channels: two for 2D displacement, and one for thickness. 
 For thickness, we use a ReLU activation to guarantee non-negative outputs, while for the 2D real-valued displacement output, we do not use any non-linearity.
 
\paragraph{Differentiable vector renderer.} 
Given the predicted displacement $\bd_i$ for each vector path $\bc_i$, new vector paths are formed as $\bc_i'=\bc_i +\bd_i$. Using the differentiable vector graphics renderer DiffVG \cite{li2020differentiable}, these new vector paths are rasterized into grayscale polylines based on predicted thickness $\bt_i$. Specifically, for each pixel in the output image, its distance to the closest point on the vector paths is computed. If it is smaller
than half the stroke thickness of the closest point, the pixel is inside the stroke's area and assigned black color; otherwise it is marked as white. The strokes are rendered 
in a  $768\times768$ raster image $\bI_b$ with anti-aliasing provided by the differentiable renderer. The resulting image is grayscale, lacking texture (see Figure \ref{fig:architecture}).

\subsection{Stroke Texture}
\label{sec:texture_module}

The final module predicts texture for all strokes. Texture may vary according to depth and underlying shape features, e.g., an artist may use darker strokes for strong shape protrusions, and lighter strokes for lower-curvature regions. As a result, we condition the texture prediction not only  on the raster drawing $\bI_b$ representing  the generated grayscale strokes, but also  the    shape representations used as input to the surface geometry module of Section \ref{sec:geometry_model}.
    Specifically, we  formulate  texture prediction as a 2D\ image translation problem.  The input to our image translation module are  the first eight channels of the view-based features $\bV$ (Section \ref{sec:geometry_model}) concatenated with the raster drawing $\bI_b$ channel-wise, resulting in a $768\times768\times9$ map $\bU$. This map is  translated into a RGB  image  $\bI =g(\bU;\bw_3)$ where $g$ is a ResNet-based fully convolutional network \cite{Johnson2016Perceptual} with four residual blocks, and $\bw_3$ are  parameters learned during training. 

As an optional post-processing step, to incorporate
the predicted texture into our editable vector graphics
representation, we convert the predicted RGB colors into a per stroke texture map. Specifically, each stroke is  parameterized by a 2D u-v map, whose coordinates are used  as a look-up table to access the texture map for each stroke. The color of each pixel in the stroke's texture map is determined by the \  RGB color of the corresponding pixel in the translated image $\bI$.

\vspace{3mm}

%% file: tex/training.tex
\label{sec:training}

\begin{figure*}
\begin{center}
\includegraphics[width=\linewidth]{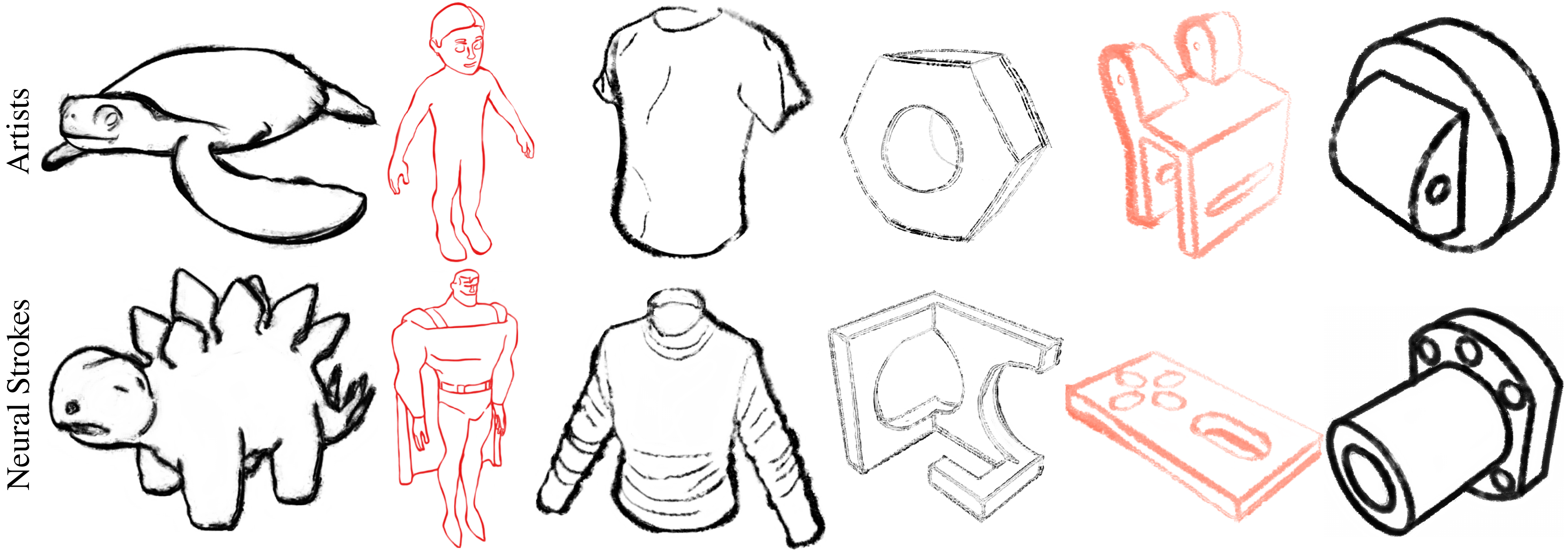}
\end{center}
\vspace{-7mm}
   \caption{A gallery of our results. \emph{Top:} artist-drawn training drawings. \emph{Bottom:} drawings from Neural Strokes.}
\vspace{-3mm}   
\label{fig:gallery}
\end{figure*}

In order to train a model, we gather drawings made by artists based on rendered line drawings. 
Due to the difficulties in creating multiple drawings in a consistent style, our training procedure is designed to work with a single training example alone. 
The goal of our training procedure is to learn the parameters $\bw=\{\bw_1, \bw_2, \bw_3\}$ of our surface geometry module, path geometry module, and stroke texture module described in Section \ref{sec:method}. 
 
\paragraph{Obtaining an artist's drawing.}
We provide an artist the feature curves $\bC$ for shape, produced from a 3D shape using the procedure in Section \ref{sec:curve_extraction}. The artist is asked to produce a drawing on a digital tablet, using the provided curves as a reference. They are specifically instructed not to trace the feature curves, so that we can capture the artist's natural tendency to deform curve shape and thickness. Given the input training drawing $\hat{\bI}$, a binary mask $\hat{\bI}_b$ is extracted by assigning black for pixels containing the artist’s strokes, and white for background. To smooth out discontinuities, anti-aliasing is applied to the mask, in the same manner as in the vector renderer, making it ``soft'' i.e., a grayscale image. 

Note that, although we have paired drawings, i.e., input 3D geometry and a drawing, our method is not fully supervised, because the drawing is provided in raster format; we do not know the stroke thickness and displacement in the drawings. This makes our data collection more flexible, allowing different data sources and allowing artists to use their favorite drawing tools.

\paragraph{Losses.}
Training a network from a single drawing is prone to overfitting. To avoid this problem, the core idea of our training procedure is to crop several random patches from the artist's drawing capturing strokes at different locations and scales.
Each of
the sampled patches is treated as a separate training instance. We use only patches that contain strokes. During training, we sample patches on the fly. For each patch, we randomly choose a crop size $c$ from a set of scales $\{64 \times 64, 128\times128, 192\times192, 256\times256\}$, crop all images and input feature maps accordingly.

We use four terms in our loss function. First, we evaluate the cropped grayscale image $\bI_b^c$ produced by the vector graphics renderer, as compared to the cropped reference soft mask $\hat{\bI}_b^c$, using $L_1$ loss:

\begin{equation} 
\mathcal{L}_{b} = ||\bI_b^c - \hat{\bI}_b^c||_1
\end{equation}
Using the above loss alone, we found that the network sometimes end up generating implausible self-intersecting  and noisy strokes. To handle this problem, we add a shape regularization term on the predicted  displacements: 
\begin{equation}
    \mathcal{L}_{s} = \frac{1}{N^c}\sum_{i=1}^{N^c} \frac{1}{(M_i^c-1)} \sum_{j=1}^{M_i^c-1} ||\bd_{i,j} - {\bd_{i, j + 1}}||^2
\end{equation}
where $N^c$ is the number of vector paths in the cropped patch and $M_i^c$ is the number of points on the path $i$.

We use $L_1$ loss in RGB space for texture, 
comparing a crop $\bI^c$ from our predicted drawing and the corresponding crop $\hat{\bI}^c$ from the artist's drawing $\hat{\bI}$:
\begin{equation}
    \mathcal{L}_t = ||\bI^c - \hat{\bI}^c||_1
\end{equation}

Finally, we use an adversarial loss to encourage the output patches to
be visually similar to random patches from the artist's drawing. To this end, we add a discriminator $\D$ during training that is trained in parallel with the stroke texture module. Architecturally, the discriminator $\D$ is identical to a $70 \times 70$ PatchGAN \cite{isola2017image} with instance normalization, and it employs a standard
LSGAN \cite{mao2017least} discriminator loss. The output patches of our model are taken as \textit{fake}, and random patches from the artist's drawing are
taken as \textit{real}. The patches are always selected to contain stroke pixels. We add the adversarial loss below to our stroke texture module by encouraging output patches to be classified as \textit{real} by the discriminator $\D$:
\begin{equation}
    \mathcal{L}_a = (\D(\bI^c)  - 1 )^2
\end{equation}

\begin{figure*}[t!]
\begin{center}
\includegraphics[width=\linewidth]{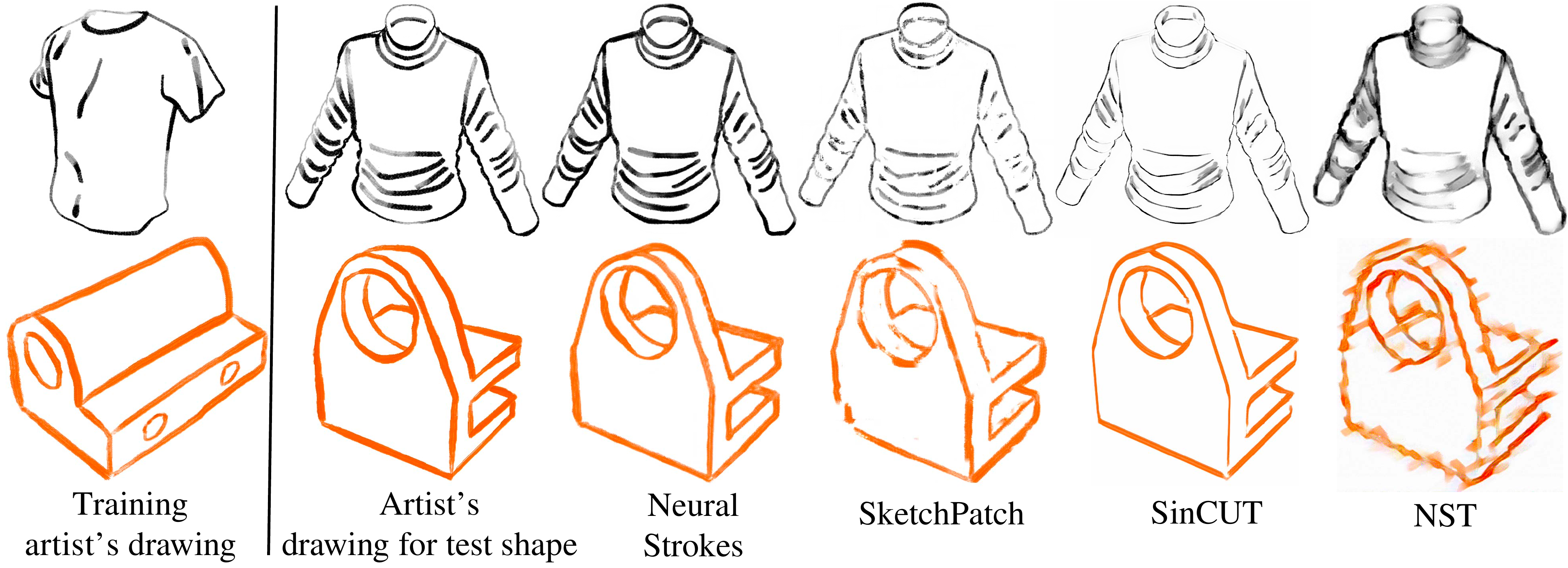}
\end{center}
\vspace{-6mm}
   \caption{Comparisons with other methods. \emph{Left to right:} training artist's drawing, artist's drawing for test shape, Neural Strokes, SketchPatch, SinCUT, NST result.
   Where possible, we retrained the other methods to incorporate the same geometry features present in the 3D shape as in our method. 
   Our method produces strokes having more similar texture, intensity and thickness variation to the artist's drawing compared to other methods, which seem to miss the above style aspects.}
\label{fig:comparisons}
\vspace{-3mm}  
\end{figure*}

\paragraph{Implementation details.} We  train the surface geometry and path geometry modules using $\lambda_b \mathcal{L}_b + \lambda_s \mathcal{L}_s$, and train the stroke texture module with $\lambda_t \mathcal{L}_t+ \lambda_a \mathcal{L}_a$. The hyperparameters are set to the default values: $\lambda_b=1,\lambda_s=0.02,\lambda_t=1,\lambda_a=1$.
For all three modules, we used the Adam optimizer \cite{kingma2014adam} with learning rate set to $0.0002$ and batch size $16$. 

%% file: tex/result.tex
We evaluated our method both qualitatively and quantitatively. Below we discuss our dataset, evaluation metrics,
comparisons with baseline methods, and our ablation study.

\paragraph{Dataset.} To create our dataset, we collected $48$ 3D shapes from an online repository (TurboSquid \cite{turbosquid})
and the ABC dataset \cite{koch2019abc}, spanning several categories, including
animals, humanoids, human body parts, clothes, and mechanical parts. All the 3D shapes are oriented and normalized so that the longest bounding box dimension is equal to $1$. A camera position is selected for each 3D shape under the constraint that it is aligned with the upright axis and points towards the centroid of the mesh. For each 3D shape and selected camera, a set of 2D geometric curves is extracted automatically using the geometry branch of Neural Contours~\cite{Liu:2020:NC}.

We hired $12$ professional artists via UpWork to stylize the 2D plain line drawings of the 3D shapes.
Each artist drew with $2$ to $4$  different styles, resulting in $31$ total styles. For each style, the artist stylized $4$ plain drawings representing $4$ different 3D shapes, resulting in a total of $124$ drawings. 
Specifically, through a web questionnaire, an artist is shown each of the four 3D shapes rendered in grayscale color using a frontal view and two side views. We also provide the artist with the plain line drawing for each of the $4$ shapes. The artist is explicitly instructed to use a textured brush, change the shape and vary the thickness of their strokes as they see fit to achieve their preferred style, and be 
stylistically consistent for all $4$ drawings.  Since our model is trained in a single image setting, for each style, we randomly select one drawing as training and keep the other three for testing and evaluation.

\paragraph{Qualitative Results.} 
 Results of our method are shown in Figure \ref{fig:teaser} and Figure \ref{fig:gallery}. As shown in the leftmost image in Figure \ref{fig:gallery}, our method accurately transfers variations in stroke thickness from the turtle to the dinosaur, giving thicker strokes to low-curvature regions; strokes are also thicker on right-facing parts of the surface. The method also transfers the charcoal-like stroke texture. In the second example, our method accurately transfers the thin strokes, with stroke thickness often thicker around convex bulges.
 
\paragraph{Evaluation metrics.} 
To perform our evaluation, we compare synthesized test drawings with ones drawn by artists. We use the following metrics for evaluation: (1) \textbf{LPIPS} Learned Perceptual Image Patch Similarity \cite{zhang2018perceptual} defined as a weighted $L_2$ distance between learned deep features of images. The measure has been demonstrated to correlate well with human perceptual similarity \cite{zhang2018perceptual}. We report LPIPS averaged over all test cases in our set across all $31$ styles. (2) \textbf{FID} Frechet Inception Distance \cite{heusel2017gans}. FID measures the distance between two set of images in terms of statistics on deep image features. In our evaluation, the set of images includes all the synthesized line drawings of our testing set, and we compare it with the set of artists' drawings.

\paragraph{Comparison methods.}  
We compare our method, Neural Strokes, with several raster image stylization approaches that attempt to transfer the style from a single example image. \textbf{(1) \emph{SketchPatch}} \cite{fish2020sketchpatch} is a paired image-to-image translation model that, like ours, operates on a patch level. During training, SketchPatch takes as input the plain line drawing patches and generates stylized line drawing patches. 
For a fair comparison, we also condition the SketchPatch model on the input shape representations of the surface geometry module (Section \ref{sec:geometry_model}) by using them as additional input channels.
We also experimented with using one SketchPatch model for stroke geometry prediction and another SketchPatch model for stroke texture prediction; yet, results did not improve (see appendix). Thus, we show here the results from training a single SketchPatch model for both. \textbf{(2) \emph{SinCUT}}  \cite{cut} is an unpaired image-to-image translation model designed to be trained from a single image. During training of SinCUT, random crops from the training drawing are used as training instances, as in our method. We also condition the SinCUT model on the input shape representations of the surface geometry module (Section \ref{sec:geometry_model}) for a fair comparison. \textbf{(3) \emph{NST}} \cite{gatys2016image} performs artistic style transfer by jointly minimizing a content loss and a style loss. Given a training drawing and a testing shape, we use the test plain line drawing as content image and the artist's training drawing as style image. \textbf{(4) B{\'e}nard \etal} \cite{Benard:2013} performs non-parametric line drawing stylization by copying pixel values from the artist's stylized drawing to corresponding pixels in the synthesized line drawing. The correspondence is optimized by PatchMatch \cite{barnes2009patchmatch}, maximizing patch-level similarity between the reference and synthesized drawings. 
As pointed out by B{\'e}nard \etal \cite{Benard:2013}, their ``parameters are style specific;'' i.e., one has to tune the parameters for each style. For purposes of comparison, we manually tuned their parameters to obtain the best results.

\begin{table}
\begin{center}
\begin{tabular}{|c|c|c|}
\hline
Method & LPIPS $\downarrow$ & FID $\downarrow$ \\
\hline\hline
\emph{SketchPatch} & 0.1104 & 83.60  \\
\emph{SinCUT} & 0.1195 & 95.74  \\
B{\'e}nard \etal \cite{Benard:2013} & 0.1618 & 181.36  \\
\emph{NST} & 0.2782 & 155.79  \\
\cline{1-3}
\emph{Neural Strokes}& \textbf{0.0956} & \textbf{62.40}   \\
\hline
\end{tabular}
\end{center}
\vspace{-6mm}
\caption{Numerical comparisons with other methods.}
\vspace{-1mm}
\label{table:comparison_table}
\end{table}

\begin{figure}[t]
\begin{center}
\includegraphics[width=\linewidth]{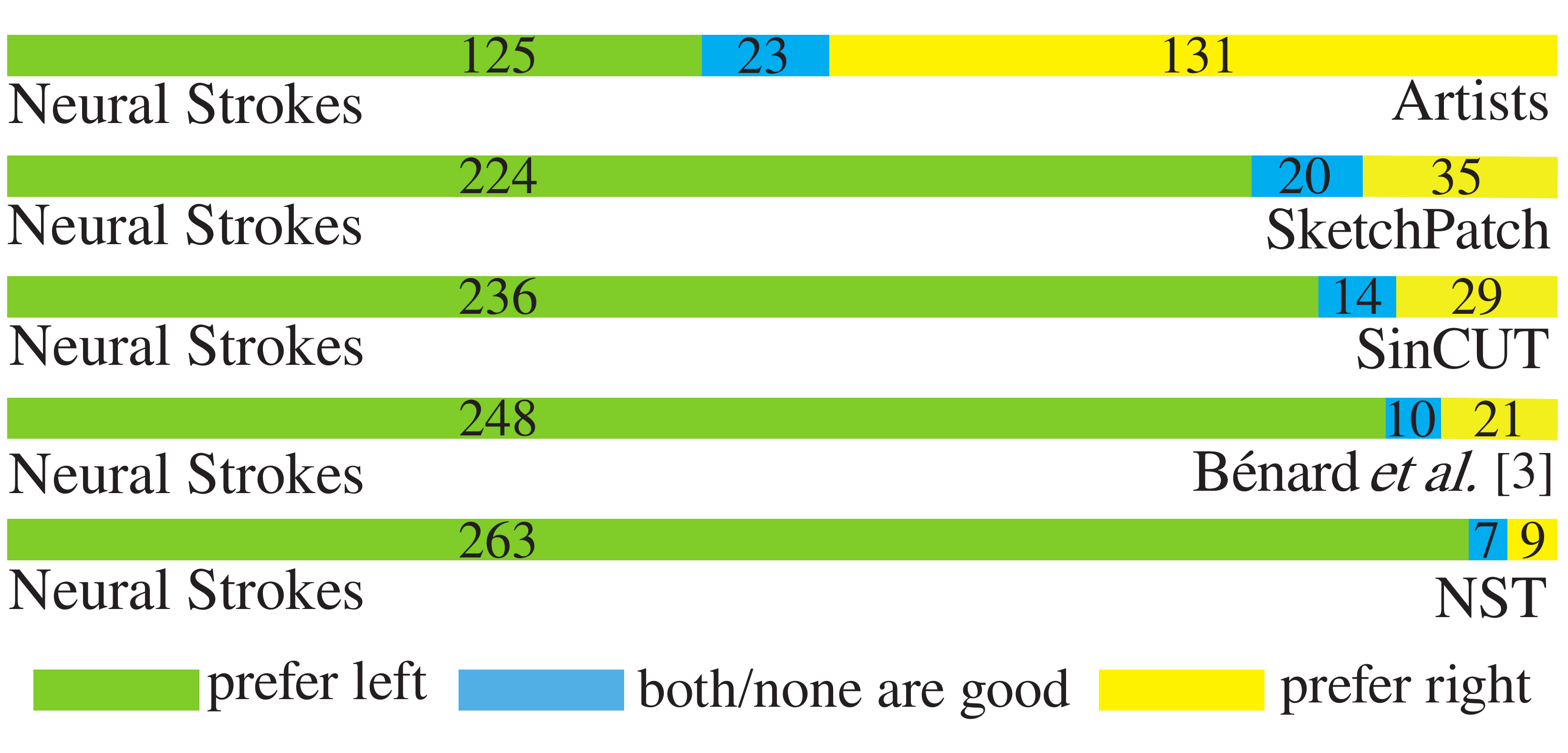}
\end{center}
\vspace{-7mm} 
   \caption{User study voting results.}
\vspace{-3mm}    
\label{fig:user_study}
\end{figure}

\paragraph{Results.} 
Table \ref{table:comparison_table} reports the evaluation measures for Neural Strokes and other competing methods. Based on
the results, Neural Strokes outperforms all competing methods in terms of both LPIPS and FID. 
Figure \ref{fig:comparisons} shows characteristic comparisons with competing methods (see also the
appendix for additional comparisons).
We also include the artists' drawings for the test shapes. 
 Since other methods do not take explicit advantage of the input geometry, except in terms of the feature maps that we provided, they often introduce gaps in strokes, or blur the stroke entirely. In the top row, observe that other methods do not accurately transfer stroke thicknesses from the example. Our method produces more precise stylized strokes, with fewer artifacts, agreeing with the artists' corresponding styles in terms of stroke thickness, shape, and texture.

\begin{table}
\begin{center}
\begin{tabular}{|c|c|c|}
\hline
Method & LPIPS $\downarrow$ & FID $\downarrow$ \\
\hline\hline
\emph{No strokes} & 0.1082 & 75.65  \\
\emph{No curve features} & 0.1006 & 67.74  \\
\emph{No surface features} & 0.1022 & 72.86  \\
\emph{No multi-scale crops} & 0.1056 & 73.70  \\
\emph{No regularization} & 0.1107 & 71.73  \\
\cline{1-3}
\emph{Neural Strokes}& \textbf{0.0956} & \textbf{62.40}   \\
\hline
\end{tabular}
\end{center}
\vspace{-6mm}
\caption{Ablation study.}
\label{table:ablation_table}
\vspace{-3mm}
\end{table}

\paragraph{User study.} 
We also conducted an Amazon MTurk study as an additional perceptual evaluation. Each questionnaire page showed participants the stylized artist's drawing for the training shape, along with a randomly ordered pair of drawings: one synthetic drawing from our method, and another from a different algorithm or from the same artist and style. We asked participants which drawing best mimicked the style of training drawing. Participants could pick either drawing,  specify ``none'' or ''both'' drawings mimicked the training drawing equally well. We asked questions twice in a random order to verify participants' reliability. We had $93$ reliable participants (see appendix for details and example questionnaires). 
Figure \ref{fig:user_study} summarizes the number of votes for the above options. The study shows that our method receives the most votes for better stylization compared to other methods, nearly seven times as many as the best alternative (SketchPatch). Moreover, our method receives similar number of votes with the  artists' drawings. This indicates that our stylized drawings are comparable to artists' drawings.

\begin{figure}[t]
\begin{center}
\includegraphics[width=\linewidth]{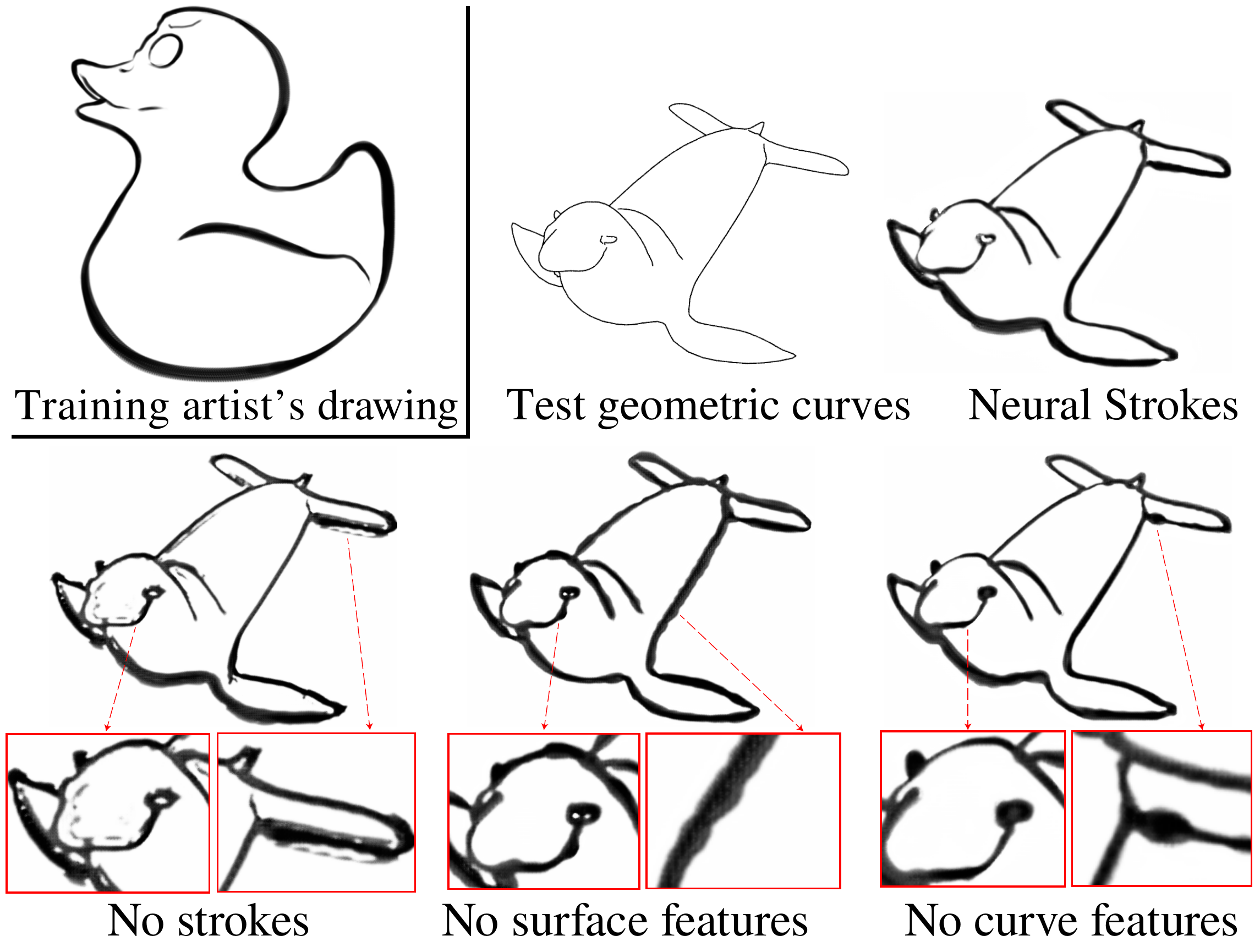}
\end{center}
\vspace{-3mm} 
   \caption{Comparisons with variants of our method. Removing features from our method result in noisy, incoherent strokes deviating from the training drawing style.}
\vspace{-1mm}    
\label{fig:ablation}
\end{figure}

\paragraph{Ablation study.}  
We also compare with the following reduced variants of our method. \textbf{(1) \emph{No strokes:}} in this reduced variant, we remove the vector stroke representation from our method and use only the surface geometry module to predict the stroke geometry as a raster image. Specifically, the surface geometry module (a 2D fully convolutional network) takes the map $\bV$ as input and produces the raster drawing $\bI_b$ directly without the use of the differentiable vector graphics renderer. Since this reduced variant does not predict thickness or displacement, we remove the displacement regularization $\mathcal{L}_s$ and only use $\mathcal{L}_b$ loss for the training of surface geometry module.  \textbf{(2) \emph{No curve features:}} we remove the raw curve features from our path geometry module. Specifically, we use the surface geometry module to produce a $768 \times 768 \times 3$ map, where each pixel contains the thickness and 2D displacement prediction. Then the stroke attributes predictions are propagated to the geometric curves directly without the use of raw curve features and our 1D CNN. \textbf{(3) \emph{No surface features:}} we exclude the 3D shape features from our path geometry module by removing the $768 \times 768 \times 8$ surface geometry renderings from the input of the surface geometry module. We also remove them from the stroke texture module. \textbf{(4) \emph{No multi-scale crops:}} during training, instead of randomly choosing a crop size from a set of scales, we use a fixed crop size $128\times128$ in this variant. \textbf{(5) \emph{No regularization:}} we remove the displacement regularization $\mathcal{L}_s$ in this variant. For all these variants, the training and architecture of stroke texture module remain the same unless specified.
Table \ref{table:ablation_table} reports the evaluation measures for Neural Strokes and the abovementioned reduced variants. The reduced variants result in worse performance. 
Figure \ref{fig:ablation} shows characteristic comparisons with the reduced variants. We observe degraded results, especially in the case of ``No strokes'' where several broken, noisy strokes appear. In the case of ``No surface features'' and ``No curve features'', we observe incoherent strokes with unnatural thickness variation and deformation.

\paragraph{Intermediate results.} 

\begin{figure}[t]
\begin{center}
\includegraphics[width=\linewidth]{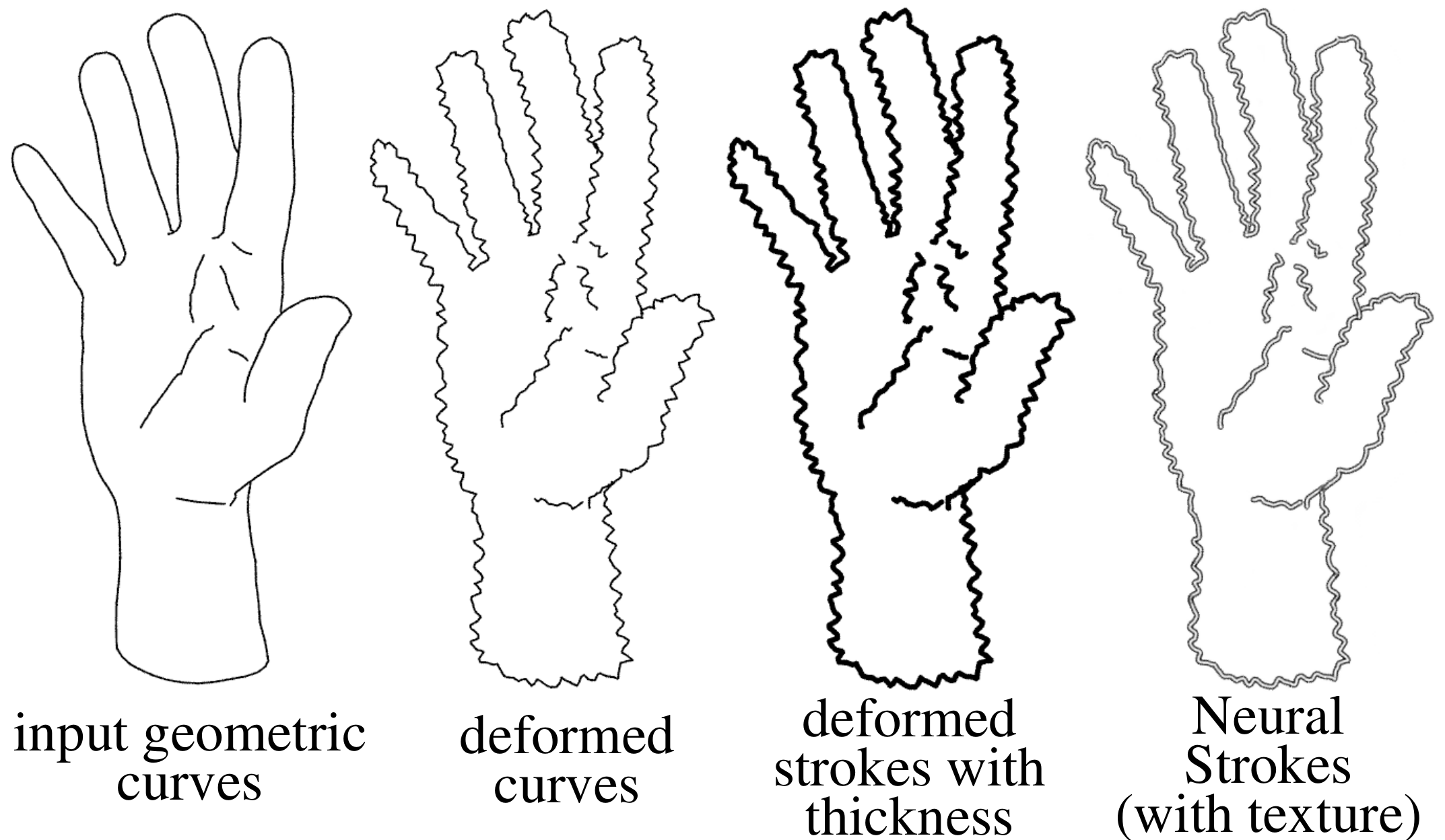}
\end{center}
\vspace{-5mm}
   \caption{
   \emph{Left to right:}
   Input geometric curves of a hand shape, deformed curves with predicted displacement from the path geometry module, strokes with predicted displacement and thickness from the path geometry module, final output textured strokes.
   }
\vspace{-3mm}      
\label{fig:intermediate}
\end{figure}

Unlike purely raster image based methods that produce strokes as pixel values, Neural Strokes predict stroke attributes (thickness, displacement, texture) in intermediate stages that can be visualized separately. 
Figure \ref{fig:intermediate} shows intermediate results of our method.
 
\begin{figure}[t]
\begin{center}
\includegraphics[width=\linewidth]{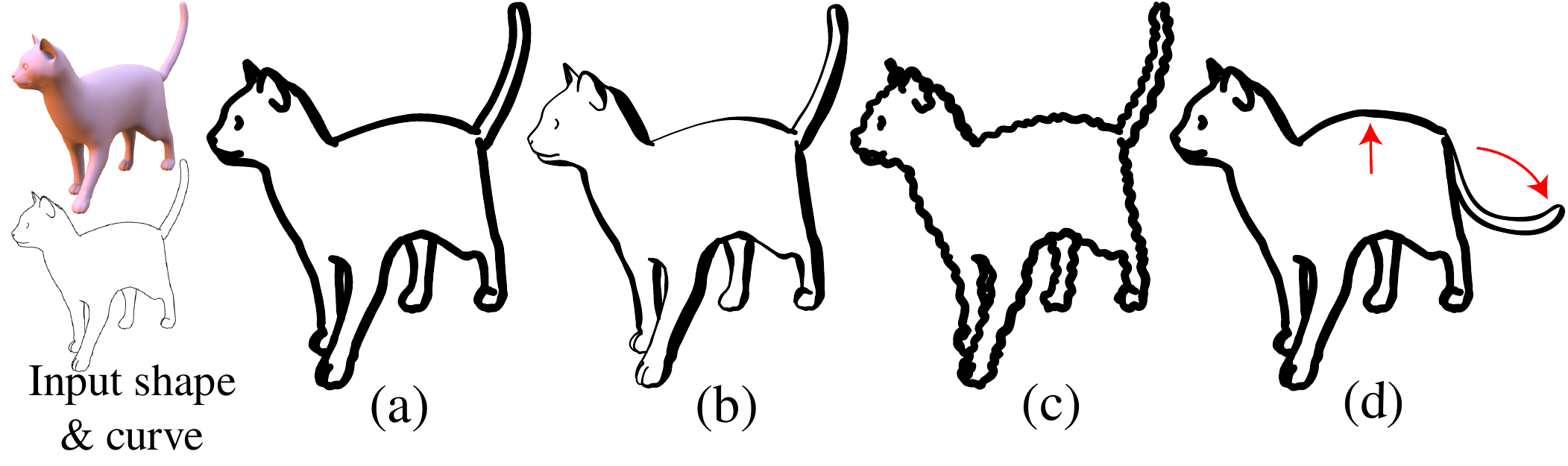}
\end{center}
\vspace{-3mm}
   \caption{Given our output strokes (a) of a cat shape, we show three editing operations: (b) rescale thickness, (c) add wiggliness, (d) move control points of strokes.}
\vspace{-1mm}      
\label{fig:editing}
\end{figure}

\paragraph{Vector Graphics editing.}
Since our method is able to output the stylized drawing in a vector representation, one can easily edit the strokes in vector graphics editing applications. In Figure \ref{fig:editing}, we show three examples of vector editing operations on our output strokes: rescaling thickness, adding wiggliness, and move control points.

%% file: tex/conclusion.tex
\begin{figure}[t]
\begin{center}
\includegraphics[width=\linewidth]{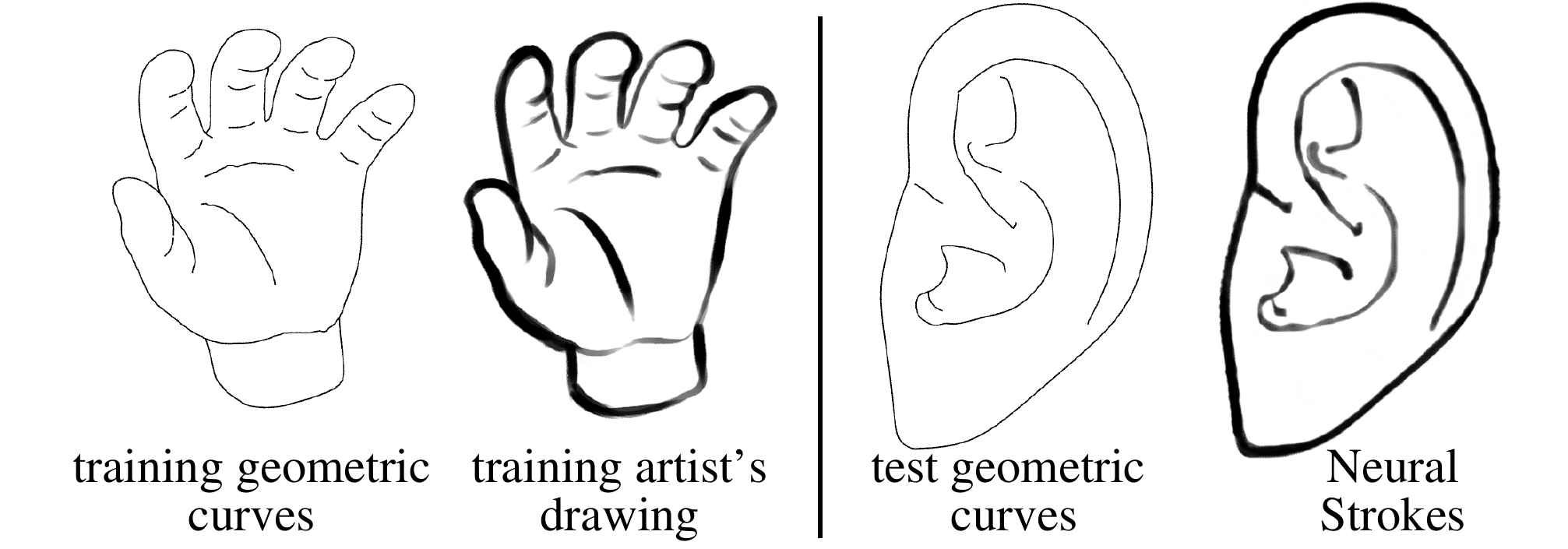}
\end{center}
\vspace{-5mm}
   \caption{Example of a less successful style transfer case.}
\vspace{-3mm}
\label{fig:failure_case}
\end{figure}

We presented a method that learns to stylize line drawings for 3D models by predicting stroke thickness, displacement and texture. The model is trained from a single raster drawing and produces output strokes in a vector graphics format. Our experiments demonstrate that our
method significantly improves over existing image-based stylization methods and that our generated drawings are comparable to artists' drawing. There are still avenues for further improvements. An artist may sometimes vary the style within the same drawing, make random choices, or the style might be uncorrelated with any of the features we use. In this case, our result may not reproduce well such stylistic choices (Figure \ref{fig:failure_case}).
In addition, when there is a large mismatch between input geometric curves and training drawing, the network may fail to reproduce correctly the stroke thickness and displacement. Learning to predict the correspondence between feature curves and the training drawing could help dealing with this issue. Learning to transfer style for other types of drawings from a single or few examples, such as hatching illustrations and cartoons,
would also be another interesting research direction. 

\paragraph{Acknowledgements.} 
This research is partially funded by NSF (CHS-1617333) and Adobe. We thank Jonathan Eisenmann and Shayan Hoshyari for helpful discussions. We also thank  the artists who contributed to our dataset.